\documentclass[10pt,twocolumn,letterpaper]{article}

\usepackage{cvpr}              % To produce the CAMERA-READY version

\usepackage{graphicx}
\usepackage{booktabs}
\usepackage{balance}
\usepackage{array}
\usepackage{makecell}
\usepackage{xcolor}

\usepackage{dingbat}
\usepackage{multirow}
\usepackage{tabularx}
\usepackage{colortbl}
\usepackage{pifont}

\usepackage{mathtools}
\usepackage{subcaption}  % 提供 subfigure 环境（现代做法）
\usepackage{wrapfig}   % for wrapfigure environment
\usepackage{amsmath}
\DeclareMathOperator*{\argmin}{arg\,min}

\definecolor{cvprblue}{rgb}{0.21,0.49,0.74}
\usepackage[pagebackref,breaklinks,colorlinks,allcolors=cvprblue]{hyperref}

\def\paperID{22815} % *** Enter the Paper ID here
\def\confName{CVPR}
\def\confYear{2026}

\title{Beyond Fixed Formulas: Data-Driven Linear Predictor\\ for Efficient Diffusion Models}

\author{
    Zhirong Shen$^{1,2}$\thanks{Equal contribution} \quad
    Rui Huang$^{1,2}$\footnotemark[1] \quad
    Jiacheng Liu$^{1,3}$ \quad
    Chang Zou$^{1,2}$ \quad
    Peiliang Cai$^{1}$ \quad
    Shikang Zheng$^{1}$ \\
    Zhengyi Shi$^{4}$ \quad
    Liang Feng$^{5}$ \quad
    Linfeng Zhang$^{1}$\thanks{Corresponding author}
    \\[0.3cm] % 增加名字与机构之间的间距
    $^1$Shanghai Jiao Tong University \quad
    $^2$University of Electronic Science and Technology of China \\
    $^3$Shandong University \quad
    $^4$Xiamen University \quad
    $^5$Fudan University
    \\[0.1cm] % 增加机构与邮箱之间的间距
    {\tt\small \{2024080907011, huang\_rui\}@std.uestc.edu.cn} \quad
    {\tt\small zhanglinfeng@sjtu.edu.cn}
}
\begin{document}
\maketitle
% \begin{abstract}
% {The iterative inference process of diffusion transformers presents a significant computational bottleneck. Although existing predictive caching methods employ complex formulas, they are constrained by their inability to be genuinely differentiated, thus resorting to finite difference approximations for derivatives. Mathematically, these methods are essentially linear combinations with fixed coefficients. These pre-determined coefficients are neither optimal nor adaptable across different models. Concurrently, our in-depth analysis of historical features has revealed a critical phenomenon: features from different time steps exhibit strong linear correlation. This creates a favorable condition for directly optimizing the predictive coefficients through a data-driven approach. Based on this finding, we propose the $L^2P$ framework, which automatically learns the optimal linear combination coefficients via gradient descent on a small dataset. This approach completely diverges from the traditional path of optimization through the discovery of different formulas. Furthermore, the $L^2P$ framework has an extremely low training cost, achieving convergence in just 20 seconds. Experiments demonstrate that on FLUX.1-dev, $L^2P$ achieves a 4.55$\times$ speedup with a PSNR of 31.459, maintaining extremely high fidelity to the original image. This performance surpasses the metrics of Taylorseer at a 2.27$\times$ speedup and significantly outperforms existing caching methods, thereby offering a novel solution for efficient diffusion model inference.}
% \end{abstract}

\begin{abstract}
Diffusion Transformers (DiTs) have achieved state-of-the-art image and video generation performance, but sampling remains expensive due to repeated transformer forward passes over many timesteps. Feature caching offers a training-free way to accelerate inference by reusing or forecasting hidden representations, yet recent forecasting-based methods derive their coefficients from hand-crafted formulas (e.g., Taylor expansion), which ultimately reduce to fixed linear combinations of a few historical features. Such fixed coefficients are suboptimal and fragile under aggressive skipping. In this paper, we first show that existing forecasting-based caching methods can be unified in a common linear form, and then analyze DiT feature trajectories, finding that for most denoising steps the current feature can be reconstructed from past features with projection fidelity above 0.95, indicating that accurate linear prediction is feasible. Motivated by this, we propose $L^2P$ (Learnable Linear Predictor), a simple data-driven caching framework that replaces hand-designed coefficients with learnable per-timestep weights trained on a small set of cached trajectories using a mean-squared error loss, converging in about 20 seconds on a single GPU. Extensive experiments on state-of-the-art DiTs demonstrate that $L^2P$ consistently outperforms existing caching baselines: on FLUX.1-dev, $L^2P$ achieves a 4.55$\times$ FLOPs reduction and 4.15$\times$ latency speedup with a PSNR of 31.459, and on Qwen-Image and Qwen-Image-Lightning, it maintains high visual fidelity even under up to 7.18$\times$ acceleration, where prior methods suffer from noticeable quality degradation. These results show that learning linear predictors is a practical and effective alternative to designing increasingly complex forecasting formulas for efficient diffusion model inference. The implementation code is available at https://github.com/Aredstone/L2P-Cache.
\end{abstract}
    
\section{Introduction}
\label{sec:intro}

Diffusion models~\cite{DM,StableDiffusion,blattmann2023SVD} have emerged as the dominant paradigm in the field of generative AI, delivering state-of-the-art performance on tasks of image and video synthesis. The integration of the Transformer architecture has given rise to diffusion transformers~\cite{DiT}, further enhancing the quality and scalability of generated visual content. A direct consequence of this enhanced capability is a substantial increase in computational cost, which impedes their application in scenarios demanding real-time performance. This computational bottleneck has motivated the development of multiple acceleration techniques~\cite{ma2024deepcache,li2023FasterDiffusion,chen2024delta-dit, liu2025survey}.

One significant training-free acceleration method is feature caching, which leverages the temporal coherence between adjacent denoising steps. The conventional "cache-and-reuse" paradigm stores features from computed time steps to be directly reused in subsequent time steps~\cite{selvaraju2024fora,zou2024accelerating}. However, while this strategy proves effective for small step intervals, it confronts a fundamental challenge: feature similarity diminishes as the time interval expands, leading to error accumulation and a decline in generation quality under high acceleration ratios. To overcome this limitation, a new paradigm is emerging: predicting future features based on past computations, rather than statically reusing them. These prediction-based methods model the temporal evolution of features~\cite{liuTaylorSeer2025}, enabling them to maintain accuracy over longer intervals and thereby achieve more substantial acceleration without sacrificing generation fidelity.

We deeply analyzed the mathematical essence of existing feature prediction methods, revealing that their core lies in the linear combination of historical features. Specifically, although such methods employ formulas like Taylor expansion and Hermite interpolation for prediction, their process of approximating derivatives through finite differences is mathematically equivalent to applying a set of fixed coefficients, determined entirely by the expansion order and step size, to a linearly weighted summation of a finite number of historical features. This finding indicates that the expressive capability of such predictive methods is fundamentally constrained by the optimality of the linear combination coefficients. To further validate the efficacy of the linear framework, we assessed the adequacy of linear representation by evaluating the projected relative residual. Experimental results demonstrate that, for the majority of time steps, the feature at the current time step can be linearly expressed by the features of the preceding steps with an error of less than one percent. This further theoretically substantiates the viability of the linear prediction paradigm.

Based on these observations, we propose $L^2P$, a novel data-driven and learnable method. The core idea of $L^2P$ is to replace the formula-derived, prior weights of traditional methods with learnable coefficients. This data-driven approach models the temporal evolution across different time steps, significantly enhancing the accuracy and robustness of long-term prediction. Concurrently, our method can be trained on a small dataset of only 50 images and achieves convergence in just 20 seconds, maintaining an extremely low training overhead. Experiments demonstrate that $L^2P$ maintains excellent generation quality and consistency with images generated by the original model, even across different models and under high acceleration ratios, surpassing all existing cache methods. 

Our contributions are as follows:

\begin{itemize}
    \item We systematically analyzed the mathematical foundation of existing feature prediction methods, revealing their nature as a fixed linear combination of historical features. We also validated the efficacy of the linear prediction scheme through empirical studies of projected residuals.
    \item We propose $L^2P$, a learnable linear predictor. It replaces traditional fixed weights with adaptive coefficients, thereby achieving more precise modeling of different models and feature evolution dynamics.
    \item Extensive experiments demonstrate that $L^2P$ maintains exceptional generation quality across various high-acceleration-ratio scenarios, while requiring only minimal training data and an extremely short training time. Its performance significantly surpasses all existing acceleration methods.
\end{itemize}

\section{Related Works}
\label{sec:rel}

The significant computational overhead of Diffusion Transformers (DiT)~\cite{ronneberger2015unet, DiT, huang2025harmonica, huang2026acceleratingdiffusionmodeltraining} stems from two main aspects: first, the extensive computation required by the Transformer module itself when processing high-resolution inputs, and second, the dozens of sequential iterative denoising steps necessary to ensure generation quality~\cite{yang2025cogvideox,li_hunyuan-dit_2024,kong2024hunyuanvideo,wan_wan_2025}. Consequently, existing acceleration research primarily revolves around these two bottlenecks.

\subsection{Acceleration Methods Requiring Model Retraining}

The first category reduces computational cost by modifying the model or compressing the sampling trajectory, but requires expensive retraining.

\noindent \textbf{Model Architecture Simplification and Compression.}
These methods act directly on the denoising network itself. Structural pruning~\cite{structural_pruning_diffusion, zhu2024dipgo} attempts to remove redundant weights or modules from the model. Another path is Token Reduction~\cite{zhang2024tokenpruningcachingbetter}, which dynamically identifies and discards (pruning) or merges (merging) image patch tokens that contribute less to the final generation during inference, thereby significantly reducing the computational~\cite{yuan2024ditfastattn} burden of the self-attention layers~\cite{saghatchian2025cached}. Although these methods can effectively reduce FLOPs, they require modifications and fine-tuning of the original model, and they risk performance degradation due to over-compression.

\noindent \textbf{Trajectory Compression.}
The goal of these methods is to distill or ``compress'' the original, lengthy diffusion path (e.g., 50 steps) into a new path with far fewer steps. For example, Consistency Models learn consistency across the trajectory through specialized training, enabling the model to jump directly from any timestep to the final clear image, achieving one-step~\cite{song2023consistency} or few-step~\cite{yan2024perflow} generation. Flow Matching~\cite{refitiedflow} constructs a more direct and stable sampling trajectory by learning a deterministic transformation path between noise and data. These methods are highly efficient in theory, but their common drawback is the need for expensive, full or partial, retraining of the diffusion model itself.

\subsection{Training-Free Inference-Time Acceleration}

In contrast, the second path accelerates unmodified diffusion models at inference time, avoiding high training costs.

\noindent \textbf{Advanced Sampling Schedulers.}
This line of research aims to design more efficient numerical solvers to use larger discrete step sizes while maintaining precision~\cite{ma2024l2c}. This includes DDIM~\cite{songDDIM}, and the more recent DPM-Solver~\cite{lu2022dpm, lu2022dpm++} series. They use high-order ODE solvers to safely increase the discretization step size. These training-free methods reduce total steps but still require a complete forward pass per step.

\textbf{Feature Caching.} This training-free approach accelerates inference by skipping computations at certain timesteps, leveraging the temporal coherence of features during diffusion. The field has evolved through two paradigms: (1) \textit{``Cache-then-Reuse''} ~\cite{ma2024deepcache,selvaraju2024fora,zou2024DuCa} cache features at timestep $t$ and directly reuse them for subsequent skipped steps. However, feature similarity drops sharply as the time interval increases, causing severe error accumulation at high acceleration ratios. (2) \textit{``Cache-then-Forecast''} methods, pioneered by TaylorSeer~\cite{liuTaylorSeer2025}, treat feature evolution as a time series and use Taylor series expansion to predict future features~\cite{Liu2025SpeCa, liu2024timestep, kahatapitiyaAdaptiveCachingFaster2024}. This inspired subsequent work like FoCa~\cite{zhengFoCa2025} and FreqCa~\cite{liu2025freqcaacceleratingdiffusionmodels}, which explore polynomial-based feature estimation for higher speedup ratios.

% \vspace{1em}
Although the \textit{``Cache-then-Forecast''} paradigm has made significant progress compared to the ``reuse'' paradigm, these predictors based on classical numerical methods (like Taylor expansion, interpolation) all share a fundamental limitation. As we analyzed in our introduction, these complex formulas are mathematically equivalent to a linear combination of historical features, and their combination coefficients are fixed priors determined entirely by the formula. This fixed, non-adaptive characteristic prevents them from achieving optimality and makes it difficult to generalize across models, which is the core bottleneck our work aims to address.
\section{Method}
\label{sec:method}

\subsection{Preliminary}

% \noindent \textbf{Diffusion Transformer (DiT). }
% DiT is a diffusion-model backbone built entirely from Transformer layers and augmented with adaptive normalization.
% Given an image at time $t$, the model operates on a sequence of patch tokens
% $\mathbf{x}_t = \{x_i\}_{i=1}^{H \times W}$, where each $x_i$ encodes one image patch.
% The network is a composition of $L$ Transformer blocks,
% $\mathcal{G} = g_L \circ \cdots \circ g_2 \circ g_1$,
% and each block combines self-attention, cross-attention, and a feed-forward subnetwork:
% \begin{equation}
% g_\ell(\cdot) = f_{\mathrm{MLP}}^{(\ell)} \left( f_{\mathrm{CA}}^{(\ell)} \left( f_{\mathrm{SA}}^{(\ell)}(\cdot) \right) \right),
% \quad \ell = 1, \dots, L.
% \end{equation}
% Here, $f_{\mathrm{SA}}^{(\ell)}$ captures intra-token dependencies within the current latent sequence, $f_{\mathrm{CA}}^{(\ell)}$ conditions on external signals (e.g., text or timestep embeddings), and $f_{\mathrm{MLP}}^{(\ell)}$ provides non-linear mixing in token space.
% Adaptive normalization layers modulate these components with timestep- and/or conditioning-dependent parameters, enabling the model to tailor its computations across the diffusion trajectory.

\noindent \textbf{Feature Caching for Diffusion Transformer.}
Temporal feature caching in DiT can be organized into two complementary paradigms: \emph{reuse-based} caching and \emph{forecasting-based} caching. 
In the reuse-based paradigm, a periodic schedule with interval $\mathcal{N}$ selects an anchor step $t$ at which all layer features are computed and stored, $\mathcal{C}(x_t^l) := \mathcal{F}(x_t^l)$ for $l \in \{0, \dots, L-1\}$. 
For the following $\mathcal{N}-1$ steps within the period, computation is skipped and features are reused,
\begin{equation}
\hat{\mathcal{F}}(x_{t+k}^l) := \mathcal{C}(x_t^l), \quad k \in \{1, \dots, \mathcal{N} - 1\},
\end{equation}
yielding an approximate FLOPs reduction of $(\mathcal{N} - 1)/\mathcal{N}$ at the cost of an accumulated temporal mismatch as $\mathcal{N}$ grows. 
In the \emph{forecasting-based} paradigm, instead of directly reusing the anchor features, the skipped-step features are \emph{predicted} from cached temporal statistics that summarize the local dynamics of features. Among forecasting-based methods, a representative approach is TaylorSeer~\citep{liuTaylorSeer2025}, which maintains the anchor feature together with its temporal finite differences up to order $m$, and estimates the feature at step $t-k$ via a truncated Taylor expansion around $t$,
\begin{equation}
\mathcal{F}_{\mathrm{pred},m}(x_{t-k}^l) = \mathcal{F}(x_t^l) + \sum_{i=1}^{m} \frac{\Delta^{i} \mathcal{F}(x_t^l)}{i! \mathcal{N}^{i}} (-k)^{i},
\end{equation}
where $\Delta^{i} \mathcal{F}(x_t^l)$ denotes the $i$-th temporal difference and the factor $\mathcal{N}^{i}$ normalizes by the period length. This Taylor-series modeling explicitly captures the short-horizon evolution of features and helps mitigate drift at larger skip intervals.

\subsection{Motivation}

\textbf{Observation 1: }\textit{The mathematical essence of prediction is a fixed linear combination.}

\textbf{Taylorseer: }In the cache-predict paradigm, methods based on Taylor expansion, represented by TaylorSeer, have achieved state-of-the-art performance. Their core idea is to use Taylor series expansion to predict features.

% An $m$-th order Taylor expansion predictor aims to use the feature $\mathcal{F}(x_{t}^{l})$ at time $t$ and its derivatives of various orders $\Delta^{i}\mathcal{F}(x_{t}^{l})$ to predict the feature $\mathcal{F}(x_{t+k}^{l})$ at time $t+k$. However, in discrete denoising steps, we cannot compute continuous derivatives. Therefore, the core of such methods lies in using finite differences to approximate these high-order derivatives.

We dissect this approximation process. According to the definition of finite difference, any $i$-th order finite difference $\Delta^{i}\mathcal{F}(x_{t}^{l})$ can be recursively expanded into a linear weighted sum of historical features:

\begin{equation}
\label{eq:diff}
\Delta^{i}\mathcal{F}(x_{t}^{l}) = \sum_{j=0}^{i}(-1)^{j}\binom{i}{j}\mathcal{F}(x_{t-jN}^{l})
\end{equation}

As shown in Eq. \ref{eq:diff}, this $i$-th order difference (i.e., the approximation of the $i$-th order derivative) is already, mathematically, a linear combination of historical features $\mathcal{F}$, determined by the binomial coefficients $\binom{i}{j}$. When we substitute this approximate value back into the prediction formula for the $m$-th order Taylor expansion:

% \begin{equation}
% \label{eq:taylor}
% \hat{\mathcal{F}}(x_{t+k}^{l}) = \mathcal{F}(x_{t}^{l}) + \sum_{i=1}^{m} \frac{\Delta^{i}\mathcal{F}(x_{t}^{l})}{i!\cdot N^{i}}(-k)^{i}
% \end{equation}

% At this point, $\hat{f}_{t+k}$ is a linear sum of the differences $\Delta^{i}\mathcal{F}$ of various orders. We substitute Eq. \ref{eq:diff} into Eq. \ref{eq:taylor}:

{\small
\begin{equation}
\label{eq:diff_further}
\begin{split}
    \hat{\mathcal{F}}(x_{t+k}^{l}) =& \mathcal{F}(x_{t}^{l})\\
    &+ \sum_{i=1}^{m} \left[ \frac{(-k)^{i}}{i!\cdot N^{i}} \right] \sum_{j=0}^{i}(-1)^{j}\binom{i}{j}\mathcal{F}(x_{t-jN}^{l})
\end{split}
\end{equation}
}

Although Eq. \ref{eq:diff_further} is formally complex, it clearly shows that the final predicted value $\hat{f}_{t+k}$ is simply computed through a series of weighted sums of historical features $\{\mathcal{F}(x_{t}^{l}), \mathcal{F}(x_{t-N}^{l}), \ldots, \mathcal{F}(x_{t-mN}^{l})\}$. By rearranging the summations above and consolidating all scalar coefficients related to a specific historical feature $\mathcal{F}(x_{t-jN}^{l})$, this formula simplifies and can be uniformly expressed in the following fixed linear combination form:

\begin{equation}
\hat{\mathcal{F}}(x_{t+k}^{l}) = \sum_{j=0}^{m} \alpha_j \cdot \mathcal{F}(x_{t-jN}^{l})
\end{equation}

Where each coefficient $\alpha_j$ is a fixed scalar value, with its expression as:

\begin{equation}
\alpha_j = \begin{cases}1 + \sum_{i=1}^{m} \frac{(-k)^{i}}{i! N^{i}} & \text{if } j = 0 \\[8pt] \sum_{i=j}^{m} \left( \frac{(-k)^{i}(-1)^{j}}{i!\cdot N^{i}} \binom{i}{j} \right) & \text{if } j \ge 1 \end{cases}
\end{equation}

The coefficients $\alpha_j$ are a fixed set of scalars, determined a priori entirely by the Taylor expansion order $m$, the step interval $N$, and the predicted step $k$.

\textbf{FoCa: }FoCa aims to stably infer future features through a ``\textit{predict-correct}'' mechanism. The core of this method is the BDF2 (second-order Backward Differentiation Formula) and a Heun corrector. FoCa relies on BDF2 to approximate the derivative. According to the definition of BDF2, any first-order derivative $\mathcal{F}^{(1)}(x_k^l)$ can be approximated by a linear combination of its historical features:

\begin{equation}\mathcal{F}^{(1)}(x_{k}^l) \approx \frac{3\mathcal{F}(x_{k}^l) - 4\mathcal{F}(x_{k-1}^l) + \mathcal{F}(x_{k-2}^l)}{2N}\end{equation}

The BDF2 predictor estimates an initial $\hat{\mathcal{F}}(x_{k+1}^l)$:
\begin{equation}
\label{eq:pred}
    \hat{\mathcal{F}}(x_{k+1}^l) = \frac{4}{3}\mathcal{F}(x_k^l) - \frac{1}{3}\mathcal{F}(x_{k-1}^l) + \frac{2N}{3}\mathcal{F}^{(1)}(x_k^l) 
\end{equation}
Substituting the BDF2 derivative approximation into Eq.~\ref{eq:pred}:

\begin{equation}
\hat{\mathcal{F}}(x_{k+1}^l) = \frac{7}{3}\mathcal{F}(x_k^l) - \frac{5}{3}\mathcal{F}(x_{k-1}^l) + \frac{1}{3}\mathcal{F}(x_{k-2}^l)
\end{equation}

Next, the prediction result is corrected:
{\small
\begin{equation}\mathcal{F}_c(x_{k+1}^l) = \mathcal{F}(x_k^l) + \frac{N}{2}[\mathcal{F}^{(1)}(x_{k-N}^l) + \mathcal{F}^{(1)}(x_{k+1}^l)]\end{equation}
}

We again use BDF2 to approximate these two derivative terms and substitute these two approximations into the corrector. Finally, by substitution, we obtain:

{\small
\begin{equation}\begin{split}
\mathcal{F}_c(x_{k+1}^l) =& \frac{3}{4}\hat{\mathcal{F}}(x_{k+1}^l) + \frac{1}{4}\mathcal{F}(x_{k-1}^l) + \frac{3}{4}\mathcal{F}(x_{k-N}^l) \\
&- \mathcal{F}(x_{k-N-1}^l) + \frac{1}{4}\mathcal{F}(x_{k-N-2}^l)
\end{split}\end{equation}
}

Expanding and combining all terms clearly reveals that the final predicted value $\mathcal{F}_{c, k+1}$ is merely a weighted sum of historical features.

This is our Observation 1: \textit{The mathematical essence of prediction is a fixed linear combination.} This inherent rigidity means that regardless of how the distribution of features $f_t$ changes across different models, the predictor uses the exact same set of $\alpha_j$ coefficients, whose values are solely determined by the formula. This partially constitutes the upper bound on the expressive power of such methods at high acceleration ratios. Other mainstream prediction methods mostly share this same property.

\textbf{Observation 2: }\textit{The current feature can be linearly represented by historical features with high precision.}

Observation 1 reveals the inherent limitation of existing cache-prediction methods: their reliance on a set of fixed, non-adaptive linear combination coefficients. This raises a critical question: \textit{Does this limitation stem from the strategy of using fixed coefficients, or from the linear combination framework itself?} If the linear framework itself is theoretically flawed with significant error, shifting to learnable coefficients would also be futile.

To validate the theoretical upper bound of the linear prediction framework, we conducted an empirical study. Instead of using the a priori coefficients $\alpha_j$ computed by methods like TaylorSeer, we turned to solving for the optimal linear approximation of the current step's final layer output feature, $\mathcal{F}(x_{t})$, within the linear subspace spanned by its historical features.

This optimal approximation, $\mathcal{F}^*(x_{t})$, is defined as the orthogonal projection of $\mathcal{F}(x_{t})$ onto the subspace $V_t$ spanned by its historical features $\{\mathcal{F}(x_{0}), \ldots, \mathcal{F}(x_{t-1})\}$:
\begin{equation}\begin{split}
    \mathcal{F}^*(x_{t}) &= \text{Proj}_{V_t}(\mathcal{F}(x_{t})) \quad \\
    \text{where} \quad V_t &= \text{span}(\mathcal{F}(x_{0}), \ldots, \mathcal{F}(x_{t-1}))
\end{split}\end{equation}
$\mathcal{F}^*(x_{t})$ represents the minimum achievable error using a linear prediction based on these historical features.

To quantify this theoretical minimum error, we calculated the projected relative residual, which is the ratio of the L2 norm of the residual vector $\mathcal{F}(x_{t}) - \mathcal{F}^*(x_{t})$ to the L2 norm of $\mathcal{F}(x_{t})$ itself:
\begin{equation}\text{Relative Residual} = \frac{\|\mathcal{F}(x_{t}) - \mathcal{F}^*(x_{t})\|_2}{\|\mathcal{F}(x_{t})\|_2}\end{equation}
The experimental results are highly encouraging: we found that, as shown in the figure, the projected relative residual is less than 5\% for about $90\%$ of the denoising time steps $t$.

This finding provides a solid empirical foundation for our method. It powerfully demonstrates that:

The linear framework is viable: The current feature $\mathcal{F}(x_{t})$ can almost always be linearly represented by its historical features with extremely high precision.
The problem lies not in the framework, but in the coefficients: The rigidity of the $\alpha_j$ in Observation 1 is the cause of the performance bottleneck.

\begin{figure}
    \centering
    \includegraphics[width=0.8\linewidth]{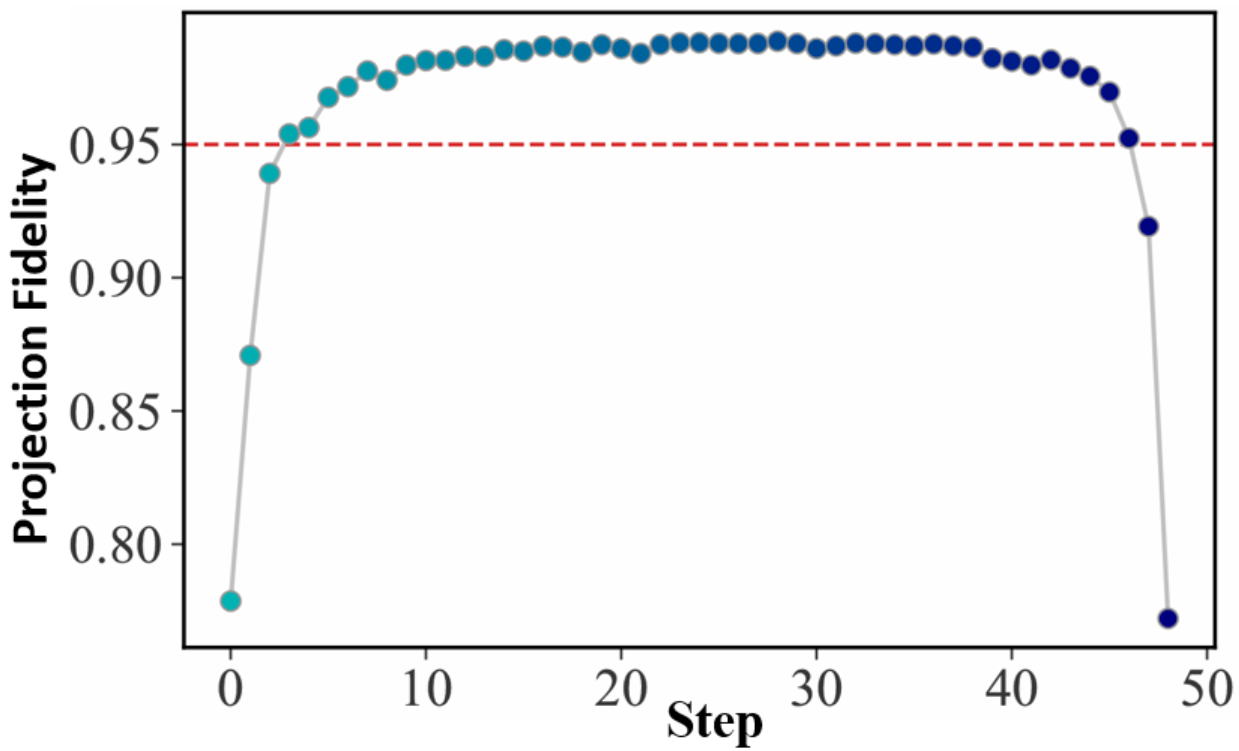}
    \vspace{-4mm}
    \caption{\textbf{Projection fidelity along a 50-step diffusion trajectory.} 
    At each step, we project the current feature onto the subspace spanned by past features and report \emph{projection fidelity}, defined as $1$ minus the relative projection residual. Most interior steps exceed $0.95$, indicating a high upper bound for linear prediction.}
    \label{fig:placeholder}
    \vspace{-6mm}
\end{figure}

\begin{figure*}
    \centering
    \includegraphics[width=\linewidth]{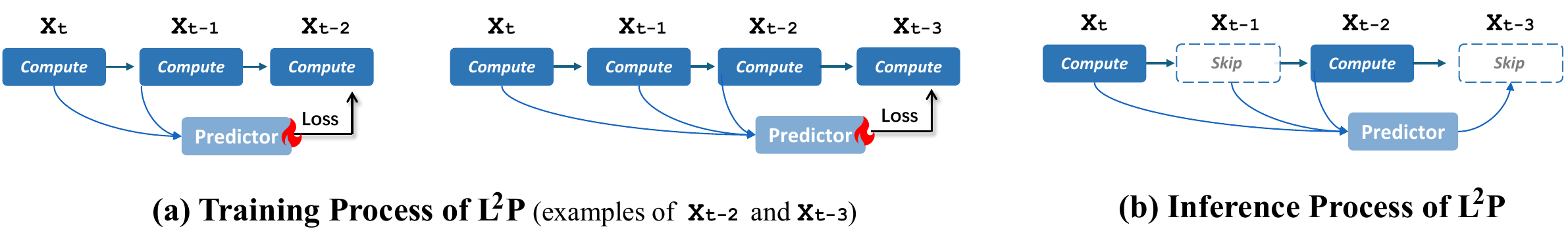}
    \vspace{-6mm}
    \caption{\textbf{Overview of the proposed $L^2P$ framework.}
    (a) During training, we still run the full diffusion trajectory and feed cached features from previous timesteps into a learnable linear predictor, which is supervised to regress the feature at the current timestep.
    (b) During inference, we reuse the cached features of the kept steps and apply the predictor to approximate features at the skipped steps, allowing us to bypass the expensive DiT forward passes for those steps.}
    \label{fig:main}
    \vspace{-15pt}
\end{figure*}

The core conclusion is that a significant performance gap exists between the performance of fixed coefficients and the potential of optimal coefficients. This points to a clear direction for our new method, $L^2P$: retain the efficient linear prediction framework, but replace the fixed, formula-derived coefficients $\alpha_j$ with data-driven, learnable coefficients $W$, thereby approximating the optimal solution $\mathcal{F}^*(x_{t})$ defined in the equation.

\subsection{The $L^2P$ Framework}

\subsubsection{Overview}
Based on the two key observations, Obs. 1 \& 2, revealed in the Motivation, we propose the $L^2P$ framework as fig. \ref{fig:main}. This framework is designed to address the inherent rigidity of existing cache-prediction methods identified in Obs. 1 and to fully leverage the high-precision linear predictability of feature evolution trajectories unveiled in Obs. 2. The core idea of $L^2P$ is to establish a principled shift: replacing the a priori, fixed coefficients $\alpha_j$ derived from mathematical formulas with data-driven, learnable linear coefficients $W$. Instead of treating this as a complex non-linear modeling task, we retain the efficiency of the linear framework while further lightweighting it. We directly fit the features of the final layer, thereby substantially reducing memory and computational overhead. $L^2P$ treats the optimal prediction $\mathcal{F}^*(x_{t})$ as a target that can be approximated through training, thereby bridging the performance gap between fixed coefficients and the theoretical optimum to achieve more precise and robust long-term prediction.

\subsubsection{Training Data Collection}
To train the learnable coefficients $W$ of the $L^2P$ predictor, we must first construct a supervised dataset $D$ containing the true feature evolution trajectories. We start with a small sample set of only 50 images. For each sample in this set, we run the standard 50-step pre-trained diffusion transformer model. During this denoising process, we capture and store the final layer feature $\mathcal{F}(x_{t})$ corresponding to each of the 50 time steps $t$. Using these stored feature sequences, we construct the training sample pairs $\mathcal{D} = \{ (X_t, Y_t) \}$, where $Y_t = \mathcal{F}(x_{t})$, and $X_t = \{\mathcal{F}(x_{0}), \ldots, \mathcal{F}(x_{t-1})\}$ represents the historical features.

\subsubsection{Model Training \& Inference}
The $L^2P$ predictor itself is designed as an extremely lightweight module. Its trainable parameters $W\in \mathbb{R}^{49 \times 49}$. The first $t$ elements of the $t$-th row $W_{t}$ of this matrix contain the linear coefficients used to predict $\mathcal{F}(x_{t})$ from the entire history $\{\mathcal{F}(x_{0}), \ldots, \mathcal{F}(x_{t-1})\}$.
The training objective is to find the optimal set of parameters $W^*$ that minimizes the prediction error on the training dataset $D$:
\begin{equation}W^* = \argmin_{W} \mathbb{E}_{(X_t, \mathcal{F}(x_{t})) \sim \mathcal{D}} \left[ \mathcal{L}(\hat{\mathcal{F}}(x_{t}), \mathcal{F}(x_{t})) \right]\end{equation}
Here, $\hat{\mathcal{F}}(x_{t})$ is the prediction from the $L^2P$ predictor for time step $t$, defined as:
\begin{equation}
\label{eq:infer}
\hat{\mathcal{F}}(x_{t}) = \sum_{j=0}^{t-1} W_{t,j} \cdot \mathcal{F}(x_{j})
\end{equation}
We use the L2 loss as the loss function $\mathcal{L}$.
We employ an efficient initialization strategy. We initialize all coefficients to 0, with the exception of $W_{t,t-1}$ (corresponding to the most recent historical step), which is initialized to 1. This initialization is mathematically equivalent to naive feature caching. Thanks to this, the training process of $L^2P$ is extremely efficient, achieving convergence in only about 20 seconds when trained on an A100. During inference, we similarly use Eq. \ref{eq:infer} to predict the final layer features from historical data.

% \subsection{Linear Combinations Seen Through High-Order Expansions}
% We demonstrated in Obs. 1 that existing difference-based prediction methods are equivalent to linear combinations. Correspondingly, from another perspective, it can be proven that any linear combination expression can always be converted into a sum of high-order differences. Therefore, our explicitly learned predictor is mathematically equivalent to an implicit high-order difference expansion:
% \begin{equation}\sum_{j=0}^{t-1} W_{t,j} \cdot \mathcal{F}(x_{j}) \equiv \sum_{i=0}^{t-1} \omega_i^* \cdot \Delta^i \mathcal{F}(x_t)\end{equation}
% Furthermore, since the $i$-th order difference $\Delta^i \mathcal{F}(x_t)$ can be seen as a standard approximation of the $i$-th order derivative $\mathcal{F}^{(i)}(x_t)$, it can be regarded as a generalized expansion formula. Traditional methods (like Taylorseer) pre-define a set of fixed coefficients from a generic formula (like Taylor expansion). In contrast, from this perspective, we have found an optimal expansion formula tailored to the current model and time step through data-driven learning. It can capture the complex feature evolution dynamics more accurately than any fixed, generic formula.

\section{Experiment}
\label{sec:exp}

\subsection{Experiment Settings}

\textbf{Experimental Settings.} We train our learnable predictor for 200 epochs with a learning rate of 0.01 using 50 LLM-generated prompts. We evaluate on \textbf{FLUX.1-dev}~\cite{flux2024}, \textbf{Qwen Image}~\cite{Wu2025QwenImageTR} (including its 8-step Lightning version), and \textbf{HunyuanVideo} ~\cite{kong2024hunyuanvideo}(see Appendix C), comparing against recent feature caching and prediction methods using their official implementations. Generation fidelity is measured by PSNR, SSIM~\cite{wang2004imagequality}, and LPIPS~\cite{zhangUnreasonableEffectivenessDeep2018}. \textit{Additional experimental details are provided in the supplementary material.}

\subsection{Main Result}
\begin{table*}[ht]
    \centering
    \caption{\textbf{Quantitative comparison in text-to-image generation} for FLUX.}
    \vspace{-3mm}
    \setlength\tabcolsep{8.0pt}
    \renewcommand{\arraystretch}{0.9} % 压缩行间距
      \normalsize
      \resizebox{0.99\textwidth}{!}{
      \begin{tabular}{l | c  c | c  c | c | c | c}
        \toprule
        {\bf Method} & {\bf Latency(s) $\downarrow$} & {\bf Speed $\uparrow$} & {\bf FLOPs(T) $\downarrow$}  & {\bf Speed $\uparrow$} & {\bf PSNR$\uparrow$} & {\bf SSIM$\uparrow$} & {\bf LPIPS$\downarrow$} \\
        \midrule
      
      $\textbf{[dev]: 50 steps}$  & {26.25}  & {1.00$\times$} & {3719.50}   & {1.00$\times$} & {-}  & {-}  & {-}      \\
      % \midrule
      % 1.24983
      % {$60\%$\textbf{ steps}}  & {10.16} & {1.50$\times$} & {2231.70} & {1.67$\times$} & {30.310} & {0.7819} & {0.2461}      \\
      % {$50\%$\textbf{ steps}}  & {8.84} & {1.41$\times$} & {1859.75} & {2.00$\times$} & {29.576} & {0.7337} & {0.3106}             \\
      % {$40\%$\textbf{ steps}}  & {7.45} & {2.05$\times$} & {1487.80} & {2.62$\times$} & {29.122} & {0.6971} & {0.3619}             \\
      % {$34\%$\textbf{ steps}}  & {6.24} & {2.45$\times$} & {1264.63} & {3.13$\times$} & {28.881} & {0.6776} & {0.3913}             \\
      \midrule
        % {$60\%$\textbf{ steps}}  & {15.91} & {1.65$\times$} & {2231.70} & {1.67$\times$} & {30.310} & {0.7819} & {0.2461}      \\
        {$50\%$\textbf{ steps}}  & {13.32} & {1.97$\times$} & {1859.75} & {2.00$\times$} & {29.576} & {0.7337} & {0.3106}\\
        $\textbf{DBCache}$ & {17.94}   & {1.46$\times$} & {1390.35}   & {2.67$\times$} & {31.208} & {0.7946} &  {0.2258} \\
        $\textbf{FORA}$ $(\mathcal{N}=3)$ & {10.33}   & {2.54$\times$} & {1320.07}   & {2.82$\times$} & {30.654} & {0.7667} &  {0.2450} \\
        $\textbf{TeaCache}$ $(l=0.6)$ & {8.95}   & {2.93$\times$} & {1412.89}   & {2.63$\times$} & {29.368} & {0.7013} &  {0.3705} \\
        $\textbf{TaylorSeer}$ $(O=2,\mathcal{N}=3)$ & {11.57}   & {2.27$\times$} & {1320.07}   & {2.82$\times$} & {30.769} & {0.7817} &  {0.2299} \\
        
      \midrule
        $\textbf{FORA}$ $(\mathcal{N}=5)${\textcolor{red}{$^{\dagger}$}} & {7.10}   & {3.69$\times$} & {893.54}   & {4.16$\times$} & {28.432} & {0.6029} &  {0.4918} \\
        $\textbf{TeaCache}$ $(l=0.8)$ & {7.05}   & {3.72$\times$} & {892.35}   & {4.17$\times$} & {28.753} & {0.6531} &  {0.4417} \\
        $\textbf{\texttt{ToCa}}$ $(\mathcal{N}=5)$ & {14.65}   & {1.79$\times$} & {1126.76}   & {3.30$\times$} & {29.328} & {0.6994} &  {0.3457} \\
        $\textbf{\texttt{DuCa}}$ $(\mathcal{N}=5)$ & {9.56}   & {2.74$\times$} & {1078.34}   & {3.45$\times$} & {29.642} & {0.7082} &  {0.3319} \\
        $\textbf{TaylorSeer}$ $(O=2,\mathcal{N}=5)$ & {8.62}   & {3.04$\times$} & {893.54}   & {4.16$\times$} & {29.328} & {0.6994} &  {0.3457} \\
        $\textbf{FoCa}$ $(\mathcal{N}=5)$ & {7.46}   & {3.51$\times$} & {893.54}   & {4.16$\times$} & {29.413} & {0.7142} &  {0.3082} \\
        
        \rowcolor{gray!20}
        $\textbf{Ours}$ $(\mathcal{N}=5)$ & {\textbf{6.32}}   & {\textbf{4.15}$\times$} & {\textbf{818.28}}  & {\textbf{4.55}$\times$} & {\textbf{31.459}} & {\textbf{0.8028}} & {\textbf{0.2147}} \\
        
      \midrule
      
        $\textbf{FORA}$ $(\mathcal{N}=7)${\textcolor{red}{$^{\dagger}$}} & {5.91}   & {4.44$\times$} & {670.44}   & {5.55$\times$} & {28.315} & {0.5870} &  {0.5409} \\
        $\textbf{TeaCache}$ $(l=1.0)${\textcolor{red}{$^{\dagger}$}} & {5.97}   & {4.39$\times$} & {743.63}   & {5.00$\times$} & {28.606} & {0.6360} &  {0.4773} \\
        $\textbf{\texttt{ToCa}}$ $(\mathcal{N}=9)${\textcolor{red}{$^{\dagger}$}} & {11.63}   & {2.25$\times$} & {784.54}   & {4.74$\times$} & {28.889} & {0.6407} &  {0.4525} \\
        $\textbf{\texttt{DuCa}}$ $(\mathcal{N}=8)${\textcolor{red}{$^{\dagger}$}} & {7.20}   & {3.64$\times$} & {676.79}   & {5.50$\times$} & {29.133} & {0.6150} &  {0.4534} \\
        $\textbf{TaylorSeer}$ $(O=2,\mathcal{N}=7)$ & {7.32}   & {3.58$\times$} & {670.44}   & {5.55$\times$} & {28.671} & {0.6237} &  {0.4542} \\
        $\textbf{FoCa}$ $(\mathcal{N}=7)$ & {6.36}   & {4.12$\times$} & {670.44}   & {5.55$\times$} & {29.193} & {0.6620} &  {0.3876} \\
        
        \rowcolor{gray!20}
        $\textbf{Ours}$ $(\mathcal{N}=7)$ & {\textbf{5.36}}   & {\textbf{4.89}$\times$} & {\textbf{669.50}}  & {\textbf{5.56}$\times$} & {\textbf{30.627}} & {\textbf{0.7524}} & {\textbf{0.2828}} \\

      \midrule

        $\textbf{FORA}$ $(\mathcal{N}=9)$ {\textcolor{red}{$^{\dagger}$}} & {5.63}   & {4.66$\times$} & {670.44}   & {5.55$\times$} & {28.315} & {0.5870} &  {0.5409} \\
        $\textbf{TeaCache}$ $(l=1.2)${\textcolor{red}{$^{\dagger}$}} & {5.11}   & {5.14$\times$} & {669.27}   & {5.56$\times$} & {28.131} & {0.4744} &  {0.6765} \\
        $\textbf{\texttt{ToCa}}$ $(\mathcal{N}=12)${\textcolor{red}{$^{\dagger}$}} & {10.45}   & {2.51$\times$} & {644.70}   & {5.77$\times$} & {28.575} & {0.5677} &  {0.5500} \\
        $\textbf{\texttt{DuCa}}$ $(\mathcal{N}=10)${\textcolor{red}{$^{\dagger}$}} & {6.46}   & {4.06$\times$} & {606.91}   & {6.13$\times$} & {28.953} & {0.5957} &  {0.4935} \\
        $\textbf{TaylorSeer}$ $(O=2,\mathcal{N}=9)${\textcolor{red}{$^{\dagger}$}} & {6.72}   & {3.90$\times$} & {596.07}   & {5.55$\times$} & {28.381} & {0.5895} &  {0.5092} \\
        $\textbf{FoCa}$ $(\mathcal{N}=8)$ & {5.88}   & {4.46$\times$} & {596.07}   & {6.24$\times$} & {29.047} & {0.6375} &  {0.4195} \\
        
        \rowcolor{gray!20}
        $\textbf{Ours}$ $(\mathcal{N}=10)$ & {\textbf{4.35}}   & {\textbf{6.03}$\times$} & {\textbf{520.72}}  & {\textbf{7.14}$\times$} & {\textbf{30.031}} & {\textbf{0.7113}} & {\textbf{0.3545}} \\
        
      \midrule
      \end{tabular}
      }
      \raggedright
    {\scriptsize
    \begin{itemize}[leftmargin=10pt,topsep=0pt]
    \item\textcolor{red}{$\dagger$} Methods exhibit significant degradation in image quality. 
    \end{itemize}
    }
      \label{table:FLUX-Metrics}
    \end{table*}

\subsubsection{FLUX.1-dev}

\begin{figure*}
    \centering
    \includegraphics[width=\linewidth]{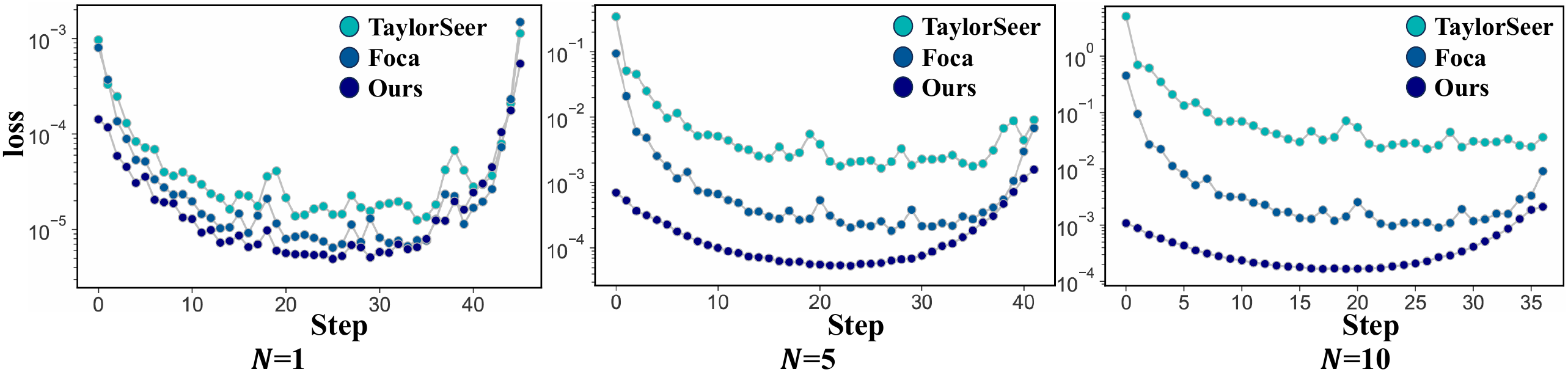}
    \vspace{-6mm}
    \caption{\textbf{MSE Loss Comparison Across Intervals.} Logarithmic comparison of MSE loss between our method, TaylorSeer, and FoCa for predictions at different step intervals ($N$ = 1, 5, 10). The results highlight the performance of each method in terms of prediction accuracy over varying distances.}
    \label{fig:placeholder}
    \vspace{-15pt}
\end{figure*}

As detailed in Table \ref{table:FLUX-Metrics}, our comprehensive comparison on the FLUX.1-dev model demonstrates that our method achieves substantial speedups while maintaining high fidelity across all acceleration levels. At $\mathcal{N}=5$, our method yields a $4.55\times$ FLOPs reduction and a PSNR of 31.459, outperforming TaylorSeer and FoCa, which show notable fidelity degradation with PSNRs of 29.328 and 29.413, respectively. This performance gap widens at higher acceleration levels. At $\mathcal{N}=7$ with a $5.56\times$ FLOPs reduction, our method maintains a PSNR of 30.627, whereas FoCa and TaylorSeer drop to 29.193 and 28.671. Under extreme acceleration at $\mathcal{N}=10$ with a $7.14\times$ FLOPs reduction, our approach uniquely maintains high-quality generation with a PSNR of 30.031, far surpassing all baselines falling below 29.1. These results underline our method's superior data efficiency and robustness at extreme acceleration levels.

\begin{figure*}
    \centering
    \includegraphics[width=\linewidth]{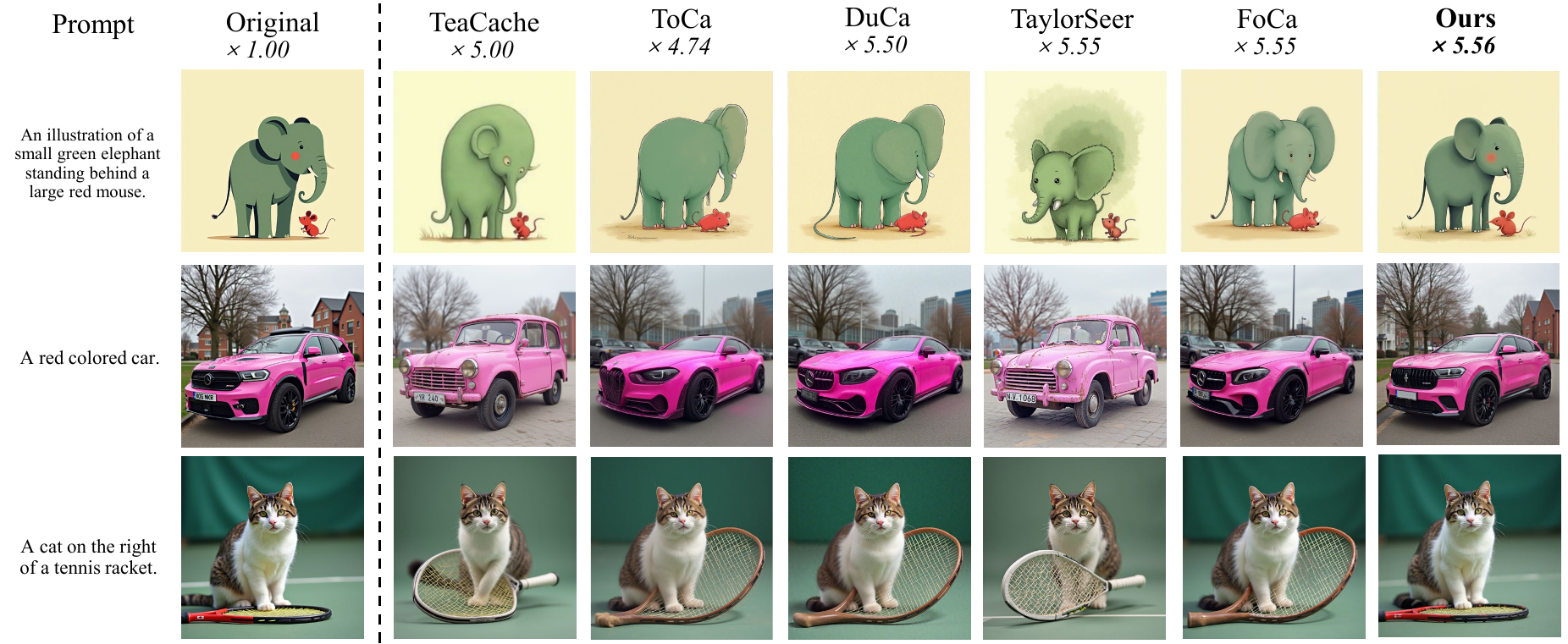}
    \vspace{-6mm}
    \caption{\textbf{Image Generation Comparison Across Methods.} Comparison of image generation results for different methods (TeaCache, ToCa, DuCa, TaylorSeer, FoCa, and Ours) based on a set of prompts. The results show the generated images for each method along with the scaling factors, highlighting how each model interprets the given prompts.}
    \vspace{-4mm}
    \label{fig:placeholder}
\end{figure*}

% 1.37 92.99
\begin{table*}[ht]
    \centering
    \caption{\textbf{Quantitative comparison in text-to-image generation} for Qwen Image.}
    \vspace{-3mm}
    \setlength\tabcolsep{8.0pt} 
    \renewcommand{\arraystretch}{0.9} % 压缩行间距
      \normalsize
      \resizebox{0.99\textwidth}{!}{
      \begin{tabular}{l | c  c | c  c | c | c | c}
        \toprule
        {\bf Method} & {\bf Latency(s) $\downarrow$} & {\bf Speed $\uparrow$} & {\bf FLOPs(T) $\downarrow$}  & {\bf Speed $\uparrow$} & {\bf PSNR$\uparrow$} & {\bf SSIM$\uparrow$} & {\bf LPIPS$\downarrow$} \\
        \midrule

\textbf{50 steps}
& 127.40 & 1.00$\times$ & 12917.56 & 1.00$\times$ & $\infty$ & 1.00 & 0.00 \\

% \textbf{$50\%$ steps}
% & 64.10 & 1.99$\times$ & 6458.78 & 2.00$\times$ & 30.54 & 0.75 & 0.28 \\

\textbf{$20\%$ steps} {\textcolor{red}{$^{\dagger}$}}
& 25.92 & 4.89$\times$ & 2583.51 & 5.00$\times$ & 28.59 & 0.61 & 0.52 \\

\midrule

\textbf{FORA}($\mathcal{N}$=4){\textcolor{red}{$^{\dagger}$}}
& 38.43 & 3.32$\times$ & 3359.99 & 3.84$\times$ & 28.66 & 0.59 & 0.51 \\

\textbf{\texttt{ToCa}}($\mathcal{N}$=8){\textcolor{red}{$^{\dagger}$}}
& 61.37 & 2.08$\times$ & 2991.34 & 4.32$\times$ & \underline{28.93} & \underline{0.63} & \underline{0.44} \\

\textbf{\texttt{DuCa}}($\mathcal{N}$=9){\textcolor{red}{$^{\dagger}$}}
& 34.73 & 3.67$\times$ & 2958.13 & 4.37$\times$ & 28.45 & 0.58 & 0.55 \\

\textbf{TaylorSeer}($\mathcal{N}$=6)
& \underline{30.75} & \underline{4.14}$\times$ & \underline{2583.97} & \underline{5.00}$\times$ & 28.58 & 0.62 & 0.46 \\

\rowcolor{gray!20}
\textbf{Ours}($\mathcal{N}$=7)
& \textbf{28.16} & \textbf{4.52}$\times$ & \textbf{2312.86} & \textbf{5.59}$\times$ & \textbf{30.62} & \textbf{0.80} & \textbf{0.21} \\

\midrule

\textbf{FORA}($\mathcal{N}$=6){\textcolor{red}{$^{\dagger}$}}
& 28.69 & 4.44$\times$ & 2326.74 & 5.55$\times$ & 28.48 & 0.55 & 0.59 \\

\textbf{\texttt{ToCa}}($\mathcal{N}$=12){\textcolor{red}{$^{\dagger}$}}
& 50.95 & 2.50$\times$ & 2406.20 & 5.37$\times$ & \underline{28.69} & \underline{0.57} & \underline{0.53} \\

\textbf{\texttt{DuCa}}($\mathcal{N}$=12){\textcolor{red}{$^{\dagger}$}}
& 28.57 & 4.46$\times$ & 2171.56 & 5.95$\times$ & 28.38 & \underline{0.57} & 0.60 \\

\textbf{TaylorSeer}($\mathcal{N}$=9){\textcolor{red}{$^{\dagger}$}}
& \underline{24.64} & \underline{5.17}$\times$ & \underline{2067.29} & \underline{6.25}$\times$ & 28.25 & 0.56 & 0.58 \\

\rowcolor{gray!20}
\textbf{Ours}($\mathcal{N}$=10)
& \textbf{23.50} & \textbf{5.42}$\times$ & \textbf{1798.89} & \textbf{7.18}$\times$ & \textbf{29.60} & \textbf{0.75} & \textbf{0.29} \\

\midrule
\textbf{Qwen-Image-Lightning-8steps}
& 17.76 & 1.00$\times$ & 2123.17 & 1.00$\times$ & $\infty$ & 1.00 & 0.00 \\

\rowcolor{gray!20}
\textbf{Ours ($\mathcal{N}$=2)}
& 12.68
& 1.40$\times$ 
& 1326.98
& 1.60$\times$
& \textbf{32.871}
& \textbf{0.8524} 
& \textbf{0.1054} \\

\rowcolor{gray!20}
\textbf{Ours ($\mathcal{N}$=3)}
& 10.91
& 1.63$\times$ 
& 1061.59
& 2.00$\times$
& \textbf{32.068} 
& \textbf{0.7988}
& \textbf{0.1535} \\
\midrule
\end{tabular}
}

\label{table:qwen-image-Metrics}
\raggedright
{\scriptsize
\begin{itemize}[leftmargin=10pt,topsep=0pt]
\item\textcolor{red}{$\dagger$} Methods exhibit significant degradation in image quality. 
\end{itemize}
}
\vspace{-6mm}
\end{table*}

\subsubsection{Qwen Image}
   \vspace{-2mm}

As detailed in Table \ref{table:qwen-image-Metrics}, validation on the Qwen Image model confirms our robust performance across acceleration levels. At $\mathcal{N}=7$, our approach achieves a $5.59\times$ FLOPs reduction, a $4.52\times$ latency speedup, and a high PSNR of 30.62. This significantly outperforms degrading baselines like ToCa at $\mathcal{N}=8$ with a PSNR of 28.93 and TaylorSeer at $\mathcal{N}=6$ with 28.58. Under extreme acceleration at $\mathcal{N}=10$ yielding a $7.18\times$ FLOPs reduction, our method uniquely maintains a PSNR of 29.60, whereas all baselines drop below 28.7. Furthermore, integrating our plug-and-play method at $\mathcal{N}=3$ into the highly-optimized 8-step Qwen-Image-Lightning model yields an additional $1.63\times$ latency speedup and $2.00\times$ FLOPs reduction, maintaining a near-lossless PSNR of 32.068 that surpasses the standard 50-step baseline.

\begin{figure}
    \centering
    \includegraphics[width=\linewidth]{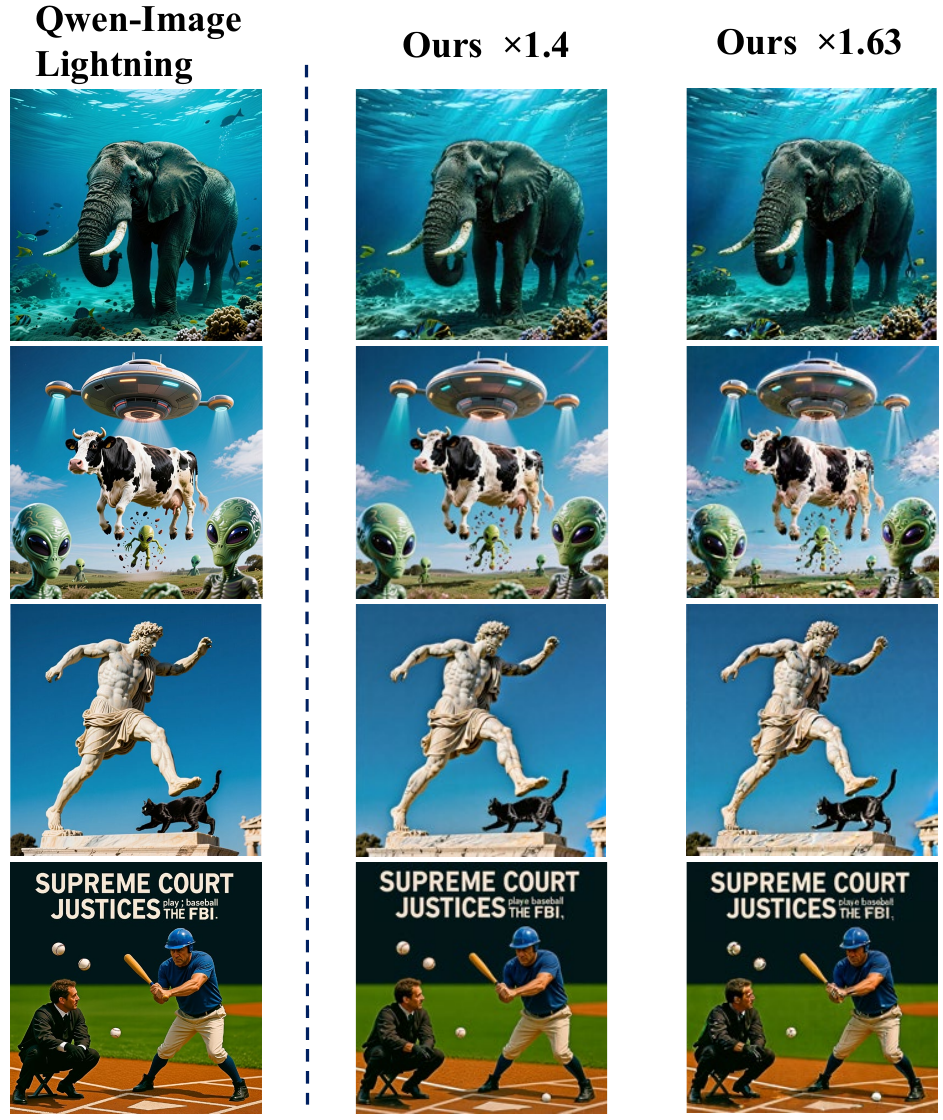}
    \vspace{-6mm}
    % \caption{\textbf{Accelerating Qwen-Image Lightning with Our Method.} Comparison of generated images using Qwen-Image Lightning with and without the addition of our method at different scaling factors (×1.4 and ×1.63). The results demonstrate how our approach accelerates Qwen-Image Lightning, providing faster and more efficient image generation while maintaining superior quality across diverse prompts.}

\caption{\textbf{Comparison of Qwen-Image Lightning} with and without our method at scaling factors $\times$1.4 and $\times$1.63, showing faster generation with maintained quality.}
    \label{fig:placeholder}
    \vspace{-8mm}
\end{figure}

\subsection{Ablation Studies}
\subsubsection{Effect of Training Sample Size}

% We first conduct an ablation study on the data efficiency of our method, specifically analyzing the impact of the number of samples used to train the learnable linear predictor. We fixed the acceleration ratio at $\mathcal{N}$=10 (corresponding to a $7.14\times$ FLOPs reduction) and then varied the number of training samples collected from the 50-step baseline from 5 to 100.
% The results are presented in Fig. \ref{fig:training}. The analysis reveals that our method is extremely data-efficient.

% \noindent \textbf{Effectiveness with Few Samples:} Even when using only 5 training samples, our method (PSNR 29.412) already significantly outperforms a strong baseline like TaylorSeer at a comparable acceleration ratio ($\mathcal{N}=9$, $5.55\times$ FLOPs reduction), which achieved a PSNR of 28.381.
% \noindent \textbf{Performance Improvement and Saturation:} As the number of samples increases from 5 to 10, there is a noticeable improvement in performance (PSNR jumps from 29.412 to 29.810, and LPIPS improves from 0.3973 to 0.3544). When the sample size is further increased to 50, the PSNR reaches its peak at 30.031.
% \noindent \textbf{Data Efficiency:} When the sample size is increased from 50 to 100, none of the metrics (PSNR 30.019, SSIM 0.7115, LPIPS 0.3531) show any further significant improvement, indicating that performance has saturated.

\begin{figure}
    \centering
    \includegraphics[width=\linewidth]{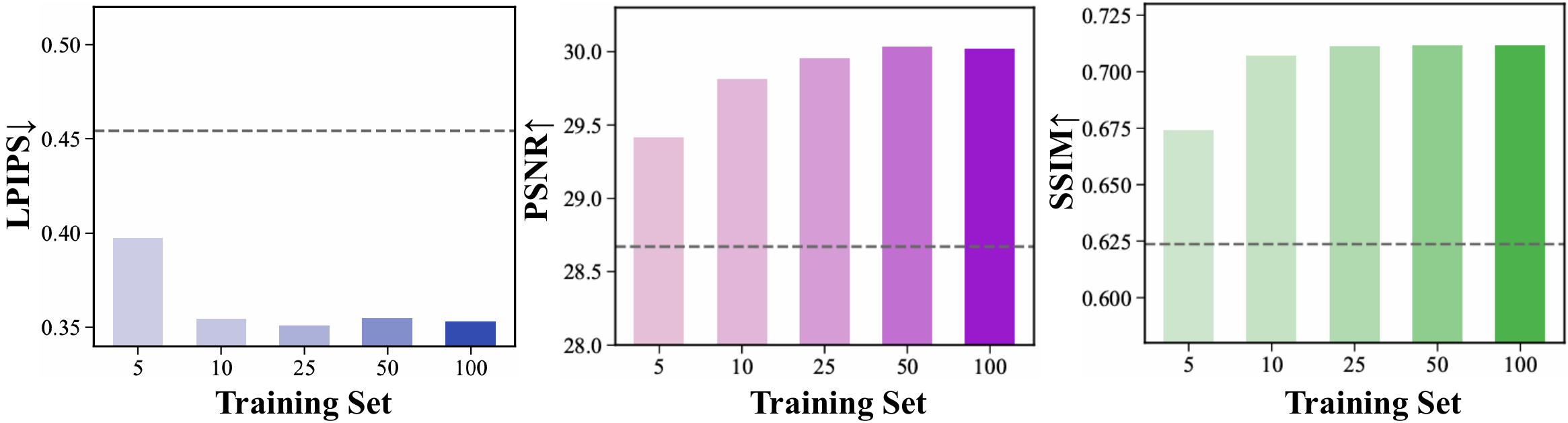}
    \vspace{-6mm}
    \caption{\textbf{Ablation Study on Data Efficiency. }Analysis of our method's performance with varying training set sizes (5, 10, 25, 50, 100) at $\mathcal{N}=10$. Results show significant improvements with few samples, with performance saturating after 50 samples.}
    \vspace{-2mm}
    \label{fig:training}
\end{figure}

As illustrated in Fig. \ref{fig:training}, we conduct an ablation study on the data efficiency of our learnable linear predictor. Fixing the acceleration ratio at $\mathcal{N}=10$ to achieve a $7.14\times$ FLOPs reduction, we varied the training samples collected from the 50-step baseline between 5 and 100.

\noindent \textbf{Effectiveness with Few Samples:} Using just 5 training samples, our method yields a PSNR of 29.412. This significantly outperforms the TaylorSeer baseline at a comparable setting of $\mathcal{N}=9$ with a $5.55\times$ FLOPs reduction, which achieves a PSNR of only 28.381.

\noindent \textbf{Performance Improvement and Saturation:} Increasing the samples from 5 to 10 delivers a noticeable performance jump, raising the PSNR to 29.810 and improving LPIPS to 0.3544. The PSNR subsequently reaches its peak at 30.031 when the sample size expands to 50.

\noindent \textbf{Data Efficiency:} Expanding the sample size further from 50 to 100 results in saturated metrics, stabilizing at a PSNR of 30.019, an SSIM of 0.7115, and an LPIPS of 0.3531 without additional significant gains.

\begin{table}[t]
\label{tab:ood}
\centering
\caption{\textbf{Ablation on Data Semantics and Feature Evolution.}}
\vspace{-3mm}
\resizebox{\columnwidth}{!}{%
\begin{tabular}{l c c c c c}
    \toprule
    \textbf{Method} & \textbf{FLOPs(T) $\downarrow$} & \textbf{Speed $\uparrow$} & \textbf{PSNR$\uparrow$} & \textbf{SSIM$\uparrow$} & \textbf{LPIPS$\downarrow$} \\
    \midrule
    \textbf{[dev]: 50 steps} & 3719.50 & 1.00$\times$ & - & - & - \\
    \midrule
    \textbf{TaylorSeer} $(O=2,\mathcal{N}=7)$ & 670.44 & 5.55$\times$ & 28.671 & 0.6237 & 0.4542 \\
    \textbf{TaylorSeer} $(O=2,\mathcal{N}=9)$ & 596.07 & 5.55$\times$ & 28.381 & 0.5895 & 0.5092 \\
    \midrule
    \textbf{Ours} $(\mathcal{N}=7)$, random & 669.50 & 5.56$\times$ & 30.627 & 0.7524 & 0.2828 \\
    \textbf{Ours} $(\mathcal{N}=10)$, random & 520.72 & 7.14$\times$ & 30.031 & 0.7113 & 0.3545 \\
    \midrule
    \textbf{Ours} $(\mathcal{N}=7)$, counterfactual & 669.50 & 5.56$\times$ & 30.707 & 0.7524 & 0.2828 \\
    \textbf{Ours} $(\mathcal{N}=10 )$, counterfactual & 520.72 & 7.14$\times$ & 30.093 & 0.7108 & 0.3518 \\
    \midrule
    \textbf{Ours} $(\mathcal{N}=7)$ , gibberish& 669.50 & 5.56$\times$ & 30.430 & 0.7517 & 0.2868 \\
    \textbf{Ours} $(\mathcal{N}=10)$, gibberish & 520.72 & 7.14$\times$ & 29.787 & 0.7099 & 0.3592 \\
    \bottomrule
\end{tabular}%
}
\label{table:Ablation-Metrics}
\vspace{-2mm}
\end{table}

This finding strongly demonstrates that our method requires only a \textbf{minimal amount} of training data (like \textbf{50} samples) to learn the optimal linear combination coefficients, showcasing its high data efficiency and practicality.

\subsubsection{Robustness to Semantic Content of Training Data}

We designed an ablation study to investigate whether our method relies on the semantic content of the training data or merely learns the dynamics of feature evolution itself. We trained three separate predictors using features collected from: 1) \textbf{Random} conventional prompts, 2) \textbf{Counterfactual} prompts describing scenarios that violate natural laws, and 3) \textbf{Gibberish} prompts composed of meaningless random characters. We then evaluated these models on the standard DrawBench benchmark, with results in Table \ref{table:Ablation-Metrics}.

The most significant finding is our method's consistent superiority over baselines regardless of training data. At $\mathcal{N}=7$ with a $5.56\times$ FLOPs reduction, even the ``gibberish'' model achieves a PSNR of 30.430, far exceeding TaylorSeer at 28.671. This advantage persists at $\mathcal{N}=10$ with a $7.14\times$ FLOPs reduction. Furthermore, semantic content shows minimal impact. The ``Counterfactual'' and ``Random'' models perform almost identically at $\mathcal{N}=7$ with PSNRs of 30.7 and 30.6, showing no dependence on factual logic. Despite a minor drop in the ``gibberish'' model to 30.4 at $\mathcal{N}=7$ and 29.8 at $\mathcal{N}=10$, it remains well above baselines, confirming our approach learns the inherent, content-agnostic linear correlation of feature evolution.

\begin{figure}
    \centering
    \vspace{-1mm}
    \includegraphics[width=\linewidth]{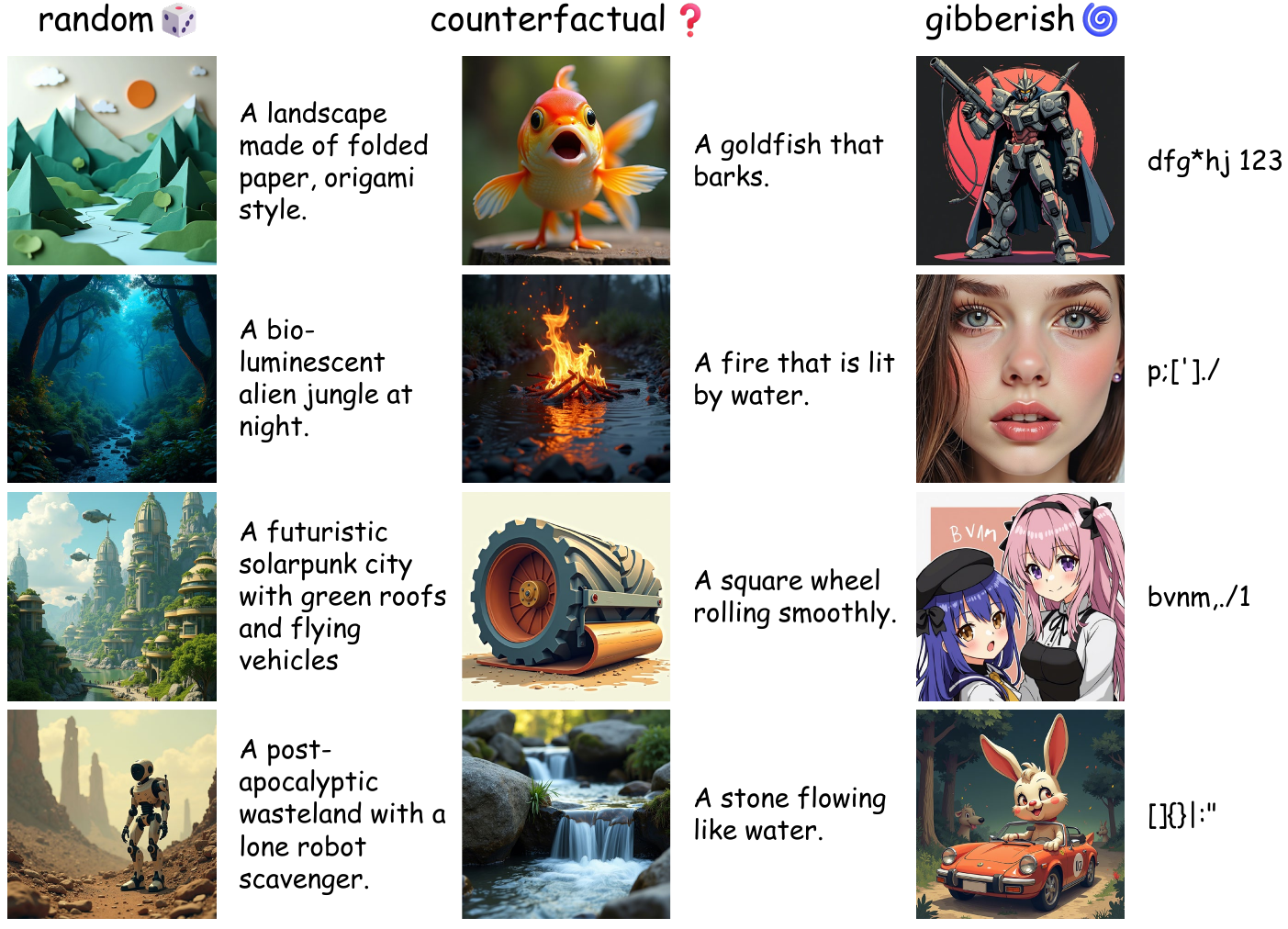}
    \caption{Illustration of Different Semantic Prompts.}
    \label{fig:ood}
\vspace{-7mm}
\end{figure}

\section{Conclusion}
We revisited feature caching for diffusion transformers from a unified linear viewpoint, showing that recent forecasting-based methods reduce to fixed-coefficient linear predictors over historical features. Our analysis of DiT trajectories revealed strong linearity, with most timesteps lying close to the span of past features, suggesting that accurate linear prediction is achievable. Based on this, we proposed \textbf{$L^2P$}, a lightweight learnable linear predictor that replaces hand-crafted coefficients with data-driven per-timestep weights and can be directly plugged into existing DiTs; experiments on FLUX.1-dev, Qwen-Image, Qwen-Image-Lightning, and HunyuanVideo show that $L^2P$ consistently outperforms prior caching-based accelerators, especially under high acceleration ratios.

\section{Acknowledgement}
This work was supported by the CCF-Tencent Rhino-Bird Funds.

{
    \small
    \bibliographystyle{ieeenat_fullname}
    \bibliography{main}
}

\clearpage
\setcounter{page}{1}
\maketitlesupplementary

\section{Experiment Settings and Evaluation}

\textbf{Implementation Details and Baselines.} We evaluate our proposed method across three advanced diffusion transformer architectures: FLUX.1-dev~\cite{flux2024} and Qwen Image~\cite{Wu2025QwenImageTR} for text-to-image generation, and Hunyuan Video~\cite{kong2024hunyuanvideo} for text-to-video tasks. The core learnable predictor was trained for 200 epochs with a learning rate of 0.01, utilizing a dataset of 50 prompts generated by an LLM. For comparative analysis, we benchmark against standard 50-step sampling baselines across all three models. Specifically, we employ the Rectified Flow sampler for FLUX.1-dev and standard samplers for Qwen Image and Hunyuan Video. Additionally, we include the Qwen-Image-Lightning-8steps~\cite{Wu2025QwenImageTR} to assess performance against accelerated distinct baselines. Activate steps were slightly adjusted via prediction residuals as hyperparameters. We further compare our approach with state-of-the-art feature caching and prediction methods using their official implementations to ensure fair comparisons.

\textbf{Evaluation Metrics.} Our evaluation framework comprehensively considers two core dimensions: acceleration efficiency and generation fidelity. For efficiency, we measure wall-clock inference latency and FLOPs to quantify computational load. We also report the speedup ratio relative to the 50-step baselines to demonstrate performance improvements. For fidelity, we assess the consistency between the accelerated outputs and the original 50-step generations. We quantify quality across all image and video results using three standard perceptual metrics: PSNR, SSIM~\cite{wang2004imagequality}, and LPIPS~\cite{zhangUnreasonableEffectivenessDeep2018}.

\section{Memory Overhead in Practice}
As described in \textbf{Sec.~3.3.2, line 352}, $L^2P$ performs prediction using \textbf{only the \textit{final-layer} features}. For HunyuanVideo at $480\times640$ resolution with 65 frames, the cached feature tensor is $(1, 20400, 64)$, incurring at most $\sim$\textbf{0.49\,GB} cache memory for 50 steps, using \textbf{$\sim$38.6$\times$} less VRAM than the window-based baseline, first-order TaylorSeer. The results on FLUX.1-dev are summarized in Table~\ref{tab:memory}.

\begin{table}[h!]
    \centering
    \caption{GPU memory usage comparison (MiB).}
    \label{tab:memory}
    % \vspace{-9pt}
    % \renewcommand\arraystretch{1.1}
    % \tabcolsep=6pt
    % 使用 resizebox 自动撑满行宽，防止超宽
    \resizebox{\linewidth}{!}{
    \begin{tabular}{l c c c c c}
        \toprule
        \textbf{Method} & w/o Cache & ToCa & TaylorSeer $(\mathcal{O}{=}2)$ & TaylorSeer $(\mathcal{O}{=}1)$ & \textbf{Ours} \\
        \midrule
        \textbf{Memory (MiB)} & 39,653 & 49,349 & 45,473 & 43,421 & \textbf{39,673} \\
        \bottomrule
    \end{tabular}
    }
    % \vspace{-15pt}
\end{table}

% ori 193.5
% interval 7 latency 41 interval 5 latency 49
% fix 20.95 25.04
\begin{table*}[htb]
\centering
\vspace{-4mm}
\caption{\centering
\textbf{Quantitative comparison of text-to-video generation} on HunyuanVideo.}
\vspace{-2mm}
\setlength\tabcolsep{5.0pt} 
\small
\resizebox{\textwidth}{!}{
\begin{tabular}{l | c c c c | c c c}
    \toprule
    {\bf Method} 
    &\multicolumn{4}{c|}{\bf Acceleration} 
    &\multicolumn{3}{c}{\textbf{Perceptual Metrics}}\rule{0pt}{2ex}\\
    \cline{2-8}
    {\bf HunyuanVideo}
    & {\bf Latency(s) $\downarrow$}
    & {\bf Speed $\uparrow$}
    & {\bf FLOPs(T) $\downarrow$}
    & {\bf Speed $\uparrow$} 
    & \textbf{PSNR\(\uparrow\)} 
    & \textbf{SSIM\(\uparrow\)} 
    & \textbf{LPIPS\(\downarrow\)}\rule{0pt}{2ex}\\ 
    \midrule

\textbf{Original: 50 steps} 
& 98.91 & {1.00}$\times$ & 29773.0 & {1.00}$\times$ & $\infty$ & 1.00 & 0.00 \\

\midrule
  
$22\%$ \textbf{steps}  
& 22.98 & {4.30}$\times$ & 6550.1 & 4.55$\times$ & 17.65 & 0.59 & 0.42 \\

\textbf{\texttt{ToCa}}~\cite{zou2024accelerating}($\mathcal{N}=5$, $\mathcal{R}=90\%$) 
& 26.12 & 3.79$\times$ & 7006.2 & 4.25$\times$ & 17.04 & 0.54 & 0.44 \\

\textbf{\texttt{DuCa}}~\cite{zou2024DuCa}($\mathcal{N}=5$, $\mathcal{R}=90\%$) 
& 23.08 & 4.29$\times$ & 6483.2 & 4.48$\times$ & 17.08 & 0.54 & 0.43\\

\textbf{TeaCache}~\cite{liu2024timestep}($l=0.4$) 
& 21.83 & 4.53$\times$ & 6550.1 & 4.55$\times$  & 18.25 & 0.61 & 0.38\\

\textbf{FORA}~\cite{selvaraju2024fora}($N=5$) 
& 22.61 & 4.37$\times$ & 5960.4 & 5.00$\times$ & 17.00 & 0.53 & 0.44 \\

\textbf{TaylorSeer}~\cite{liuTaylorSeer2025}($\mathcal{N}=5$, $O=1$) 
& 23.66 & 4.18$\times$ & 5960.4 & 5.00$\times$ & 17.29 & 0.55 & 0.42\\

% 17.2118    0.574348     0.393258

% \textbf{Speca}~\cite{Liu2025SpeCa}($\mathcal{N}_{\text{max}}=8$, $\mathcal{N}_{\text{min}}=2$) 
% & 23.22 & 4.26$\times$ & 5692.7 & 5.23$\times$ & 17.73 & 0.59 & 0.39\\

% \textbf{Clusca}($\mathcal{N}=5$, $O=1$, $K=16$) 
% & 24.35 & 4.06$\times$ & 5373.0 & 5.54$\times$ & 17.28 & 0.55 & 0.42\\

\rowcolor{gray!20}
\textbf{Ours($\mathcal{N}=7$)}
& \textbf{20.95} & \textbf{4.72}$\times$ & \textbf{5359.2} & \textbf{5.55}$\times$ & {\textbf{21.10}} & {\textbf{0.72}} & {\textbf{0.28}} \\

\bottomrule
\end{tabular}}
\label{table:HunyuanVideo-Metrics}
\end{table*}

\begin{figure*}
    \centering
    \includegraphics[width=1\linewidth]{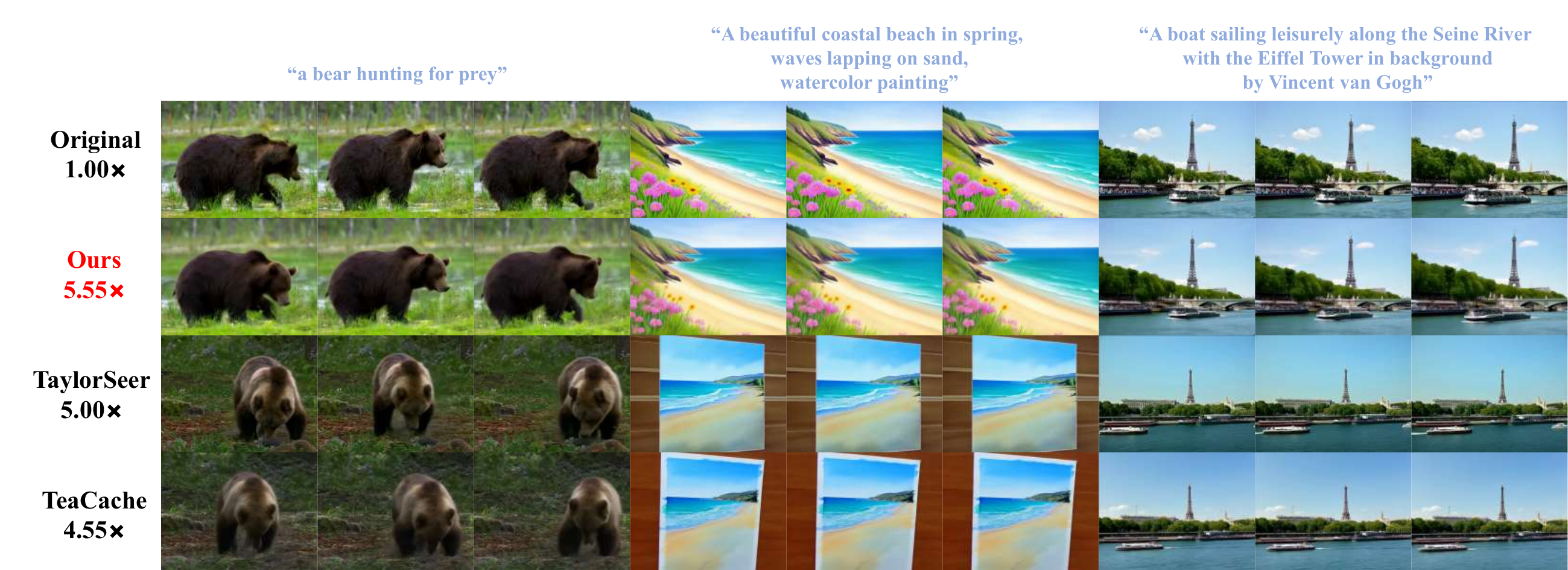}
    \caption{\textbf{Qualitative Comparison of Early, Middle, and Late Frames From Generated Videos.} On HunyuanVideo, $L^2P$ constructs
continuous dynamic process with excellent subject accuracy under high speedup ratio.}
    \label{fig:placeholder}
\end{figure*}

\section{More Results on T2V Generation}
\label{T2V}

Table \ref{table:HunyuanVideo-Metrics} reports the comprehensive performance on the HunyuanVideo benchmark. Our method demonstrates a dominant advantage across both computational efficiency and generation fidelity.

\textbf{In terms of efficiency}, our method (with $\mathcal{N}=7$) achieves the lowest inference latency of 20.95s, corresponding to a 4.72$\times$ wall-clock speedup. This significantly outperforms all competing baselines, including the recent state-of-the-art TeaCache (21.83s, 4.53$\times$) and predictive methods like TaylorSeer (23.66s, 4.18$\times$). Furthermore, our method requires the lowest computational cost (5359.2T FLOPs), validating that our lightweight linear predictor introduces negligible overhead while maximizing skipped steps.

\textbf{In terms of fidelity}, the superiority of our approach is even more pronounced. Existing acceleration methods typically struggle with the complex temporal dynamics of video, resulting in significant quality degradation (e.g., most baselines hover around 17.0 dB in PSNR). In contrast, our method maintains exceptional consistency with the original 50-step generation, achieving a PSNR of 21.10 dB. This represents a remarkable improvement of +2.85 dB over the second-best method, TeaCache. Similarly, we achieve the highest SSIM (0.72) and lowest LPIPS (0.28), proving that our data-driven predictor effectively mitigates the error accumulation that plagues fixed-coefficient predictors like TaylorSeer in long-trajectory video generation.

\section{Justification for Linear Design}

A natural question arises: \textbf{\textit{Why restrict the predictor to a linear combination? Can introducing non-linearity improve performance?}} We categorize potential non-linear enhancements into two paradigms: \textbf{(1) explicit a priori non-linear modeling}, and \textbf{(2) implicit data-driven non-linear learning.}

For the first paradigm, we conducted extensive experiments incorporating explicit non-linear priors. Our explorations included:
\begin{itemize}
    \item \textbf{Spatially-varying coefficients}: Assigning distinct linear weights to different spatial tokens rather than sharing a global scalar.
    \item \textbf{Non-linear transforms}: Introducing terms such as quadratic ($x^2$), exponential ($\exp$), logarithmic ($\log$), and radical ($\sqrt{x}$) components into the expansion.
    \item \textbf{Attention mechanisms}: Applying lightweight self-attention to capture inter-token dependencies.
\end{itemize}
However, none of these variants yielded significant performance gains. Theoretically, the underlying Stochastic Differential Equation (SDE) governing the diffusion process does not exhibit an obvious, analytically consistent prior that favors specific non-linear functional forms for feature temporal evolution.

For the second paradigm, employing deep neural networks (e.g., MLPs or LSTMs) to implicitly learn non-linear dynamics presents two critical drawbacks. First, it requires massive datasets to generalize, whereas our method works with minimal data (50 prompts). Second, and more importantly, the computational overhead of running a neural network predictor contradicts the fundamental goal of inference acceleration.

\textbf{Conclusion.} As demonstrated in \textbf{Obs. 2}, the linear subspace already captures over 95\% of the feature reconstruction fidelity for the vast majority of steps. Given that the linear framework achieves near-optimal fidelity with minimal cost, we adopt the strictly linear formulation as the optimal trade-off between efficiency and precision.

\section{Linear Combinations Seen Through High-Order Expansions}

We demonstrated in Obs. 1 that existing difference-based prediction methods are equivalentd to linear combinations. Conversely, it can be proven that any linear combination of historical features can be equivalently converted into a sum of high-order differences. Therefore, our explicitly learned predictor is mathematically isomorphic to an implicit high-order difference expansion:
\begin{equation}
\label{eq:begin}
\sum_{j=0}^{t-1} W_{t,j} \cdot \mathcal{F}(x_{j}) \equiv \sum_{k=0}^{t-1} \omega_k^* \cdot \Delta^k \mathcal{F}(x_{t-1})
\end{equation}
where $\Delta^k$ denotes the $k$-th order backward difference operator. Since the discrete difference $\Delta^k \mathcal{F}$ serves as an approximation of the $k$-th order derivative $\mathcal{F}^{(k)}$, Eq. \ref{eq:begin} can be regarded as a generalized Taylor-like expansion. While traditional methods (e.g., Taylorseer) rely on pre-defined, fixed coefficients derived from generic approximation formulas, our method learns an optimal set of coefficients tailored to the specific feature evolution dynamics. This data-driven approach allows for capturing complex temporal dependencies more accurately than rigid analytical expansions. We provide a formal proof of this equivalence below.

\paragraph{Proof of Equivalence.}
We verify that for any set of linear weights $\{W_{t,j}\}$, there exists a unique corresponding set of difference coefficients $\{\omega_k^*\}$.
First, we define the vector of historical feature values $\mathbf{f}$ and the vector of high-order backward differences $\mathbf{d}$ at step $t-1$ as:
{
\begin{align}
    \mathbf{f} &= [\mathcal{F}(x_{t-1}), \dots, \mathcal{F}(x_0)]^\top \in \mathbb{R}^{t} \\
    \mathbf{d} &= [\Delta^0 \mathcal{F}(x_{t-1}), \dots, \Delta^{t-1} \mathcal{F}(x_{t-1})]^\top \in \mathbb{R}^{t}
\end{align}
}

By definition, the $k$-th order backward difference is a linear combination of historical values determined by binomial coefficients:
\begin{equation}
    \Delta^k \mathcal{F}(x_{t-1}) = \sum_{m=0}^{k} (-1)^m \binom{k}{m} \mathcal{F}(x_{t-1-m})
\end{equation}
This relationship allows us to express the transformation in matrix form:
\begin{equation}
    \mathbf{d} = \mathbf{P} \mathbf{f}
\end{equation}
where $\mathbf{P} \in \mathbb{R}^{t \times t}$ is a lower triangular Pascal matrix with alternating signs. Its entries are given by $P_{k,m} = (-1)^m \binom{k}{m}$ for $k \ge m$, and $0$ otherwise.

Crucially, the diagonal elements of $\mathbf{P}$ are non-zero ($P_{k,k} = (-1)^k$), which implies that the determinant is non-zero:
\begin{equation}
    \det(\mathbf{P}) = \prod_{k=0}^{t-1} P_{k,k} \neq 0
\end{equation}
Thus, $\mathbf{P}$ is invertible, ensuring a bijective mapping $\mathbf{f} = \mathbf{P}^{-1} \mathbf{d}$. Substituting this into the linear predictor form (LHS of Eq. 1):
\begin{equation}
    \text{LHS} = \mathbf{W}^\top \mathbf{f} = \mathbf{W}^\top (\mathbf{P}^{-1} \mathbf{d}) = (\mathbf{W}^\top \mathbf{P}^{-1}) \mathbf{d}
\end{equation}
By defining the new coefficient vector via the transformation:
\begin{equation}
    (\boldsymbol{\omega}^*)^\top = \mathbf{W}^\top \mathbf{P}^{-1}
\end{equation}
we arrive at the high-order expansion form $\sum \omega_k^* \Delta^k \mathcal{F}(x_{t-1})$. This concludes the proof.

\end{document}